\documentclass{article}
\usepackage{spconf,amsmath,amssymb,bm}
\usepackage[dvipdfmx]{graphicx}
\usepackage{multirow}
\usepackage{xcolor}
\usepackage{comment}
\usepackage{url}

\title{Reinforcement Learning of Speech Recognition System\\
Based on Policy Gradient and Hypothesis Selection}
\name{Taku Kato, Takahiro Shinozaki\thanks{This work was supported by JSPS KAKENHI Grant Number 26280055 and 17K20001.}}
\address{Tokyo Institute of Technology, School of Engineering, Kanagawa, Japan}
\begin{document}
\ninept
\maketitle
\begin{abstract}
Speech recognition systems have achieved high recognition performance for several tasks.
However, the performance of such systems is dependent on the tremendously costly development work of preparing vast amounts of task-matched transcribed speech data for supervised training.
The key problem here is the cost of transcribing speech data.
The cost is repeatedly required to support new languages and new tasks.
Assuming broad network services for transcribing speech data for many users,
a system would become more self-sufficient and more useful
if it possessed the ability to learn from very light feedback from the users without annoying them.
In this paper, we propose a general reinforcement learning framework for speech recognition systems
based on the policy gradient method.
As a particular instance of the framework, 
we also propose a hypothesis selection-based reinforcement learning method.
The proposed framework provides a new view for several existing training and adaptation methods.
The experimental results show that the proposed method improves the recognition performance
compared to unsupervised adaptation.
% for the supervised training required to support new tasks.
%This method is based on selecting the best recognition hypothesis among two hypotheses,
%which is a binary outcome for the system for outputting two ranked hypotheses for each input utterance.
\end{abstract}
\section{Introduction}
\label{sec:intro}
Today's speech recognition systems heavily rely on supervised training using large amounts of 
task-matched training data to achieve high speech recognition performance. 
To prepare labeled speech data, a large transcription cost is required. 
This is particularly a problem for resource-limited languages. 
However, even for resource-rich languages, a significant factor that 
limits the application area of speech recognition is 
the additional transcription cost required to support new tasks that are different from the initial training condition.

When considering network applications of automatic speech recognition with many users, 
one strategy to improve the system performance without incurring 
development cost is to utilize feedback from the users while providing recognition results to them. 
Ogata et al. developed a web service called Podcastle that uses a speech recognizer to automatically transcribe speech contents in podcasts such that the users can 
read and search them~\cite{podsastle, PodCastleGoto}. 
The system includes a user interface that allows the users to correct the recognition errors word by word.
By gathering the corrected transcriptions, the speech recognition system can be 
re-estimated and improved by using any 
supervised model training or adaptation methods~\cite{MLLR, MAP, DNNadaptation_2015, Yao_SLT_2012}. 
For this system, the motivation for the users to fix the errors in the automatic transcriptions is to contribute to sharing the contents that they like. 
However, a considerable amount of effort is required to produce a correct transcription, and 
the user contribution would be limited to those contents that have enthusiastic listeners. 
If users are only asked about the recognition quality rather than the corrections of 
the errors in the transcriptions, 
and the system could utilize the scalar feedback to update the model by reinforcement learning, 
it would greatly reduce the effort required by the users. 
By reducing the effort of users, larger applications would become possible. 
%For this purpose, we propose a reinforcement learning method for DNN-HMM-based
%speech recognition systems.

Reinforcement learning is based on the common sense idea that if an 
action is followed by an improvement in the state of affairs, 
then the tendency to produce that action is strengthened~\cite{Sutton126844}. 
The two major formalizations of reinforcement learning 
are value-based methods including Q-learning approaches~\cite{Watkins1992, Rummery94on-lineq-learning, mnih-atari-2013},
and policy-based methods including policy gradient methods~\cite{Sutton:1999:PGM:3009657.3009806, pmlr-v48-mniha16}. 
In this paper, we first formulate a very general reinforcement learning framework for speech
recognition systems based on the policy gradient method.
Then, we propose a reinforcement learning method following the framework,
where the feedback is based on hypothesis selection by the users.
% cf.
% https://www.slideshare.net/sotetsukoyamada/pgq-combining-policy-gradient-and-qlearning
% https://www.cs.toronto.edu/~vmnih/
% http://citeseerx.ist.psu.edu/viewdoc/summary?doi=10.1.1.17.2539

The remainder of this paper is organized as follows. 
We first briefly review the application of reinforcement learning in speech information processing 
in Section~\ref{sec:relatedwork}, and the policy gradient method in Section~\ref{sec:policygrad}. 
We then explain our proposed method in Section~\ref{sec:proposed}
and our implementation for experiments in Section~\ref{seq:implement}. 
The experimental setup is described in Section~\ref{sec:expcond}, and the results are shown in Section~\ref{sec:results}.
Finally, the conclusions are presented in Section~\ref{sec:conclusion}.

%% \section{Reinforcement learning}
%% \label{sec:reinforcement}
%% Reinforcement learning is based on the common sense idea that if an 
%% action is followed by an improvement in the state of affairs, 
%% then the tendency to produce that action is strengthened~\cite{Sutton126844}.
%% % If it is followed by the opposite effect, then the tendency is decreased.
%% A widely used formalization of reinforcement learning is based on a Markov decision process (MDP)~\cite{Howard60}.
%% MDP is defined by a set of states of the environment,
%% a set of actions, 
%% transition probabilities of the states that depend on the action taken at a state,
%% and a reward function that depends on the state and the action.
%% A policy is a function that takes a state and returns an action.
%% The goal of a learning agent is to find an optimal policy that maximizes
%% the performance of the agent considering not only the direct reward of the current state
%% but also the future rewards.
%% In MDP, the transition probabilities may not be explicitly given to the agent.
%% In such a case, the agent learns the transition probabilities through trials.
%% However, the state of the environment is assumed to be directly observable.
%% A partially observable Markov decision process (POMDP) is an extension of MDP
%% that treats the situation where there is an ambiguity in the observation of the
%% current state.

\section{Related work}
\label{sec:relatedwork}
There have been many studies that apply reinforcement learning to
speech dialogue systems to improve dialogue control~\cite{DLu2011, PHao2015, Wang_reinforce_2010}.
For source enhancement, Koizumi et al. have proposed
a Q-learning-based method for a DNN-based system~\cite{YKoizumi2017}.
In their method, the speech enhancement performance was improved based on 
feedback from human evaluators about the perceptual quality of the enhanced speech. 
However, studies that apply reinforcement learning 
to speech recognition systems are limited, as noted in~\cite{Molina2010}.

In studies on speech recognition,
Nisida et al. have proposed a method that tunes an update coefficient $\tau$
of the MAP adaptation for GMM-HMM~\cite{Nishida2004}.
Their method used a confidence measure obtained from the result of Viterbi decoding of an utterance as the reward. 
Therefore, there was no human interaction. 
A small $\tau$ was used for speech segments with high confidence, 
and a large $\tau$ was used for segments with low confidence.
Molina et al. have proposed a two-pass decoding method 
that was also based on a confidence measure~\cite{Molina2010}.
The idea was to reinforce the phone models in the second pass if they had a high confidence value, 
whereas they were weakened if they had low confidence.
In the algorithm, the choice of the phone models in the decoding process is regarded as an action
of reinforcement learning in a broad sense. 
The confidence measure was estimated in the first pass, and it was used in the second pass by adding the value to the acoustic likelihood. 
The algorithm was for a decoding process, and the acoustic model was not updated.
These methods were based on intuitive ideas to modify the model update or decoding process 
based on the confidence measure.
However, their connections to the major formalizations of reinforcement learning methods were not explained. 
In the same sense, the two-pass unsupervised adaptation algorithms that
reject low confidence hypotheses (e.g.~\cite{Charlet2001}) 
may also be seen as a type of reinforcement learning.

\section{Policy gradient method}
\label{sec:policygrad}
As the general setup for the policy gradient method-based reinforcement learning,
a system has a set of actions and a policy function $f$
that takes a state $\bm{s}$ and returns a probability distribution $P_f\left(a|\bm{s}\right)$ of an action $a$ to take.
The policy function is parameterized by a set of parameters $\bm{\theta}$.
From $P_f\left(a|\bm{s}\right)$, an action is sampled and executed.
According to the action, the system gets a scalar reward $r_{\bm{s}}\left(a\right)$.
%Depending on the task, the reward may be immediate or delayed.

The goal of the learning is to maximize the expected reward
$\mathbb{E}\left[r_{\bm{s}}\left(a\right)\right]=\sum_a P_f\left(a|\bm{s}\right) r_{\bm{s}}\left(a\right)$
with respect to $\bm{\theta}$.
The maximization can be performed by applying the gradient ascent method.
However, the key points here are that,
while the reward $r_{\bm{s}}\left(a\right)$ can be evaluated given the choice of the action,
there may not exist an analytical functional form of the reward,
and enumerating all possible actions may not be tractable.
Therefore, we need a scheme to evaluate the gradient as follows,
which is parallel to the derivation process of the natural evolution strategy
using the $log$-trick~\cite{wierstra2011natural, hansen2003reducing}.
\begin{eqnarray}
  \nabla_{\bm{\theta}}\mathbb{E}\left[r_{\bm{s}}\left(a\right)\right|\bm{\theta}]
  &=&\nabla_{\bm{\theta}}\sum_a P_f\left(a|\bm{s}\right)r_{\bm{s}}\left(a\right) \nonumber \\
%  &=&\sum_a \nabla_{\bm{\theta}}P_f\left(a|\bm{s}\right)r_{\bm{s}}\left(a\right)\nonumber\\
  &=&\sum_a P_f\left(a|\bm{s}\right)
  \left(\frac{\nabla_{\bm{\theta}}P_f\left(a|\bm{s}\right)}{P_f\left(a|\bm{s}\right)}\right)
  r_{\bm{s}}\left(a\right)\nonumber\\
%  &=&\sum_a P_f\left(a|\bm{s}\right)\left(\nabla_{\bm{\theta}}\log P_f\left(a|\bm{s}\right)\right)r_{\bm{s}}\left(a\right)\nonumber\\
  &=&\mathbb{E}\left[r_{\bm{s}}\left(a\right)\nabla_{\bm{\theta}}\log P_f\left(a|\bm{s}\right)\right].
  \label{eq:gradexpect}
%  [f(\bm{x})|\bm{\theta}] \mid _{\bm{\theta} = \hat{\bm{\theta}}_{n-1}}\\
\end{eqnarray}
Equation \eqref{eq:gradexpect} means that $r_{\bm{s}}\left(a\right)\nabla_{\bm{\theta}}\log P_f\left(a|\bm{s}\right)$
is an unbiased estimator of the gradient $\nabla_{\bm{\theta}}\mathbb{E}\left[r_{\bm{s}}\left(a\right)\right|\bm{\theta}]$.
Given the estimate of the gradient, the parameter update formula is obtained as follows.
\begin{equation}
  \hat{\bm{\theta}} = \bm{\theta} + \epsilon r_{\bm{s}}\left(a\right)\nabla_{\bm{\theta}}\log P_f\left(a|\bm{s}\right),
  \label{eq:paramupdate}
\end{equation}
where $\epsilon \ (> 0)$ is the learning rate.
The same formulation holds when the reward is a conditional probability of $r$ given $a$.

\section{Proposed method}
\label{sec:proposed}
We assume a situation where a speech recognition
system is used to serve a vast number of general users over the Internet.
The users input speech data that they want to transcribe.
Such data would include recordings of school lectures, invited talks, presentations, and meetings.
More interactive applications, such as voice input for email, can also be the target.
The users want a reasonably good transcript quickly and easily,
and they do not have time to correct all the recognition errors word by word.
The user interface is equipped with a mechanism
that allows the users to provide a scalar evaluation score for the recognition
result as user feedback.
There are several design choices about what types of scores we expect the users to provide
intentionally or unintentionally,
but we assume that it is given in an utterance basis.

To formulate a reinforcement learning framework for statistical speech recognition systems,
we regard the whole system as a policy function 
that takes a feature sequence of an utterance as the input $\bm{s}$
and returns a probability distribution of a word sequence of recognition hypothesis as an action.
In particular, when the recognition system is based on an acoustic model $P_{AM}\left(\bm{s}|\bm{l}\right)$
and a language model $P_{LM}\left(\bm{l}\right)$,
the (unnormalized) probability distribution is given by Equation~\eqref{eq:asrpolicy}.
\begin{equation}
  \label{eq:asrpolicy}
  P\left(\bm{l}\right|\bm{s}) = \frac{P_{AM}\left(\bm{s}|\bm{l}\right)P_{LM}\left(\bm{l}\right)}{P\left(\bm{s}\right)}
  \propto P_{AM}\left(\bm{s}|\bm{l}\right)P_{LM}\left(\bm{l}\right).
\end{equation}
If we further assume that we only want to update the acoustic model
and it is a DNN-HMM, and we only want to update the DNN parameters $\bm{\theta}$ to better predict the posterior probability of HMM states,
then the gradient in Equation~\eqref{eq:paramupdate} becomes independent of the language model.
Moreover, it is further decomposed to each time frame, and becomes:
\begin{equation}
  r_{\bm{s}}\left(\bm{l}\right) \frac{\partial \log P_{AM}\left(\bm{l}_t|\bm{s}_t\right)}{\partial \bm{\theta}},
  \label{eq:dnngradreinforcement}
\end{equation}
where $\bm{s}_t$ is an acoustic feature vector at time frame $t$,
and $\bm{l}_t$ is the HMM state aligned to that frame.
Equation~\eqref{eq:dnngradreinforcement} indicates that the update formula for the
reinforcement learning of DNN-HMM using the policy gradient method is
simply a reward weighted version of normal cross-entropy based back-propagation.
The update formula satisfies the criterion of the REINFORCE algorithm
having the form shown in Equation~\eqref{eq:reinforce}~\cite{Williams1992},
\begin{equation}
  \left(r-b\right)\frac{\partial \log g(i) }{\partial \bm{\theta}},
  \label{eq:reinforce}
\end{equation}
where $r$ is the reward, $b$ is the reinforcement baseline, $g$ is the probability function
over the item $i$, and $\theta$ is a parameter set.
%Any learning algorithm of a neural network having this form is called a REINFORCE algorithm.

%% The normal supervised training of DNN-HMM is based on the gradient ascent method.
%% Let $P(s(i)|\vec{x}_t)$ be the probability of the $i$-th HMM state $s(i)$
%% estimated by a DNN acoustic model having parameter set
%% $\mat{\Theta}=\left\{\theta_1, \theta_2, \cdots, \theta_N\right\}$
%% given an input acoustic feature vector $x_t$ at time $t$,
%% where $N$ is the number of model parameters.
%% The output label $l_t$ is produced from a manually transcribed transcription,
%% which is an integer that indicates the HMM state that aligns to the time frame.
%% Based on this setup, the gradient for the $k$-th parameter $\theta_k$ is computed by Equation~\eqref{eq:dnngrad}
%% using the cross-entropy error measure~\cite{bishop2006pattern}.
%% %Equation~(\ref{})
%% \begin{equation}
%%   \frac{\partial \log P\left(s(l_t)|\vec{x}_t\right)}{\partial \theta_k}.
%%   \label{eq:dnngrad}
%% \end{equation}

%% Our basic idea is to substitute $l_t$ obtained from the manual transcription with
%% a label $h_t$ that is derived from the recognition hypothesis as in the unsupervised adaptation
%% and weight the gradient by the reward $r$, as shown in Equation~\eqref{eq:dnngradreinforcement}.
%% \begin{equation}
%%   r \frac{\partial \log P\left(s(h_t)|\vec{x}_t\right)}{\partial \theta_k}.
%%   \label{eq:dnngradreinforcement}
%% \end{equation}

If we use a confidence measure as reward
and round it to a binary value of 1.0 and 0.0, % based on a threshold,
we can now clearly state that the conventional unsupervised adaptation with the
hypothesis rejection mechanism mentioned in Section~\ref{sec:relatedwork}
is an example of the policy gradient-based reinforcement learning
if the hypothesis is obtained by the sampling. % rather than Viterbi decoding.
%since the utterances with low confidence are skipped in the model update.
%When asking the users to give a feedback, 

To utilize human feedback, the most direct measure of the recognition performance is the word accuracy.
However, asking general users to evaluate word accuracy would not be realistic.
%If we ask for such an evaluation, the scores would be inaccurate and inconsistent.
Even for users with a technical background in speech recognition,
it is time consuming to calculate. % accurate scores.
To avoid this problem, we propose a hypothesis selection-based reinforcement learning method in which
we prepare two recognition systems.
One system is the subject for the reinforcement learning, and the other is used as a rival.
For each input utterance, a recognition hypothesis is sampled from each of the systems,
and both of them are presented to the user.
Then, the user selects the better hypothesis among them.
In this case, the selection itself is the feedback to the system:
1 is the feedback when the hypothesis of the first system is selected, and
0 is the feedback otherwise.
Based on the binary reward $r$, we update the DNN using the 
weighted gradient defined in Equation~\eqref{eq:gradcompbest}.
\begin{equation}
%  \frac{1}{1+\beta}
%  \left(r_{\bm{s}}\left(\bm{l}\right) - (-\beta) \right)
%  \frac{\partial \log P_{AM}\left(\bm{l}_t)|\bm{s}_t\right)}{\partial \bm{\theta}},
  \left(1+\alpha\right)\left(r-\frac{\alpha}{1+\alpha}\right)
  \frac{\partial \log P_{AM}\left(\bm{l}_t|\bm{s}_t\right)}{\partial \bm{\theta}},
%  \frac{\partial \log P\left(s_{h1_t}|\vec{x}_t\right)}{\partial \theta_k}
%&+& \left(1+\alpha\right)\left(\left(-r\right)-\frac{-1}{1+\alpha} \right)
%  \frac{\partial \log P\left(s_{h2_t}|\vec{x}_t\right)}{\partial \theta_k}. \hspace{5mm}
  \label{eq:gradcompbest}
\end{equation}
where $\alpha$ $(0\le\ \alpha \ \le1)$ is a scalar constant. The coefficient $\left(1+\alpha\right)$ is constant and can be seen as a part of the learning rate.
Choosing $\alpha=0$ corresponds to updating the parameters only when the hypothesis is selected.

\section{Implementation with approximations}
\label{seq:implement}
To implement the proposed hypothesis selection-based reinforcement learning,
we made some approximations in our experiments.
First, we used a Viterbi decoding as in normal speech recognition systems
to find the best hypothesis rather than sampling a hypothesis from the
posterior distribution.
Second, instead of preparing a separate rival system, we used the $n$-th best hypothesis $(1<n)$
of the same system as the rival hypothesis, where $n$ is a constant.
We refer to the best hypothesis as the Candidate 1 hypothesis ($l^{(1)}$)
and the rival hypothesis as the Candidate 2 hypothesis ($l^{(2)}$).
Since both of the hypotheses come from the same model, we used both of them
in a symmetric manner in the gradient update as shown in Equation~\eqref{eq:gradcomp2best}.
\begin{eqnarray}
  && \left(1+\alpha\right)\left(r-\frac{\alpha}{1+\alpha}\right) 
  \frac{\partial \log P_{AM}\left(\bm{l^{(1)}}_t|\bm{s}_t\right)}{\partial \bm{\theta}} \nonumber \\
&+& \left(1+\alpha\right)\left(\left(-r\right)-\frac{-1}{1+\alpha} \right)
  \frac{\partial \log P_{AM}\left(\bm{l^{(2)}}_t|\bm{s}_t\right)}{\partial \bm{\theta}}. \hspace{5mm}
  \label{eq:gradcomp2best}
\end{eqnarray}
This corresponds to collecting two feedbacks for two actions at the same time.
For example, assuming $\alpha=0$, we compute the gradient using the Candidate 1 hypothesis
when it is selected (i.e. $\alpha=1$) with weight $1$,
and we use the Candidate 2 hypothesis with weight $1$ otherwise.
Third, the parameter update by the reinforcement learning was performed
based on large batches rather than an utterance by utterance update.
This is mainly for the purpose of quick implementation.

For a more rigorous implementation of the sampling from the unnormalized posterior,
beam sampling could be used~\cite{VanGael:2008:BSI:1390156.1390293}.
Another strategy of preparing a rival system would be to use
the same system from a randomly selected previous stage of update,
as in AlphaGo~\cite{Silver_2016}.
By rewriting, Equation~\eqref{eq:gradcomp2best} becomes Equation~\eqref{eq:gradcomp2best_cond}.
In this form, it can be seen that the hypothesis selection method is similar to
discriminative training~\cite{Karel_IS_2013} in that it tries to increase the difference of the likelihood of the
selected hypothesis (corresponding to correct the hypothesis) and the other hypothesis (the denominator lattice).
However, the selected hypothesis is not a reference and usually contains errors, and it is within the formulation of the expected reward.
\begin{equation}
  \left\{
    \begin{array}{ll}
    \frac{\partial \log P_{AM}\left(\bm{l^{(1)}}_t|\bm{s}_t\right)}{\partial \bm{\theta}} 
  - \alpha \frac{\partial \log P_{AM}\left(\bm{l^{(2)}}_t|\bm{s}_t\right)}{\partial \bm{\theta}} & (r=1) \\
    \frac{\partial \log P_{AM}\left(\bm{l^{(2)}}_t|\bm{s}_t\right)}{\partial \bm{\theta}}
  - \alpha \frac{\partial \log P_{AM}\left(\bm{l^{(1)}}_t|\bm{s}_t\right)}{\partial \bm{\theta}} & (r=0).
    \end{array}
  \right.
  \label{eq:gradcomp2best_cond}
\end{equation}

 \vspace{-5mm}
\begin{table}[t]
  \caption{CSJ data used for the experiments.}
  \begin{center}
        \begin{tabular}{|l|l|c|}
                \hline
%        \multicolumn{2}{|l|}{Data set}& CSJ \\\hline  
        \multirow{2}{*}{Training set}& labeled & 10 hours\\\cline{2-3}
                & unlabeled & 50 + 50 + 50 + 50 hours\\\hline
        \multicolumn{2}{|l|}{Evaluation set}& 2 hours\\\hline
                \multicolumn{2}{|l|}{Vocabulary size}& 72k words\\\hline
    \end{tabular}
         \vspace{-5mm}
  \end{center}
 \label{dataset}
\end{table}

\begin{figure}[t]
\centering
\centerline{\includegraphics[width=7.0cm]{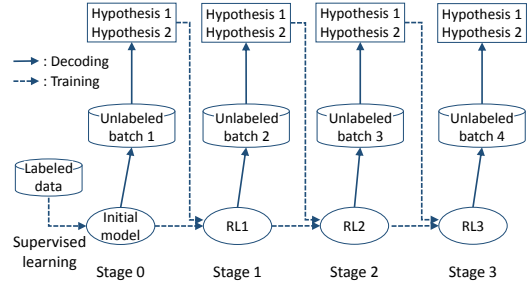}}
\vspace{-6mm}
 \caption{Reinforcement learning process.
   RLk indicates a model made by applying reinforcement learning k times.
 RLk is made at stage k and used to decode large batch \#k+1}
\label{fig:process}
% \vspace{-3mm}
\end{figure}

\section{Experimental setup}
%\ts{Please shrink from here to conclusion to two pages}
\label{sec:expcond}

%\begin{table}[t]
%  \caption{CSJ and SWBD data used for the experiments.}
%  \begin{center}
%  	\begin{tabular}{|l|l|c|c|}
%  		\hline
%        \multicolumn{2}{|l|}{}& CSJ & SWBD\\\hline
%        \multirow{2}{*}{Training set}& labeled & \multicolumn{2}{|c|}{10 hours}\\\cline{2-4}
%        	& unlabeled & \multicolumn{2}{|c|}{50 + 50 + 50 + 50 hours}\\\hline
   %     \multicolumn{2}{|l|}{Evaluation set}& 2 hours&4 hours\\\hline
%  		\multicolumn{2}{|l|}{Vocabulary size}& 72k words &30k words\\\hline
%    \end{tabular}
%  \end{center}
% \label{dataset}
%\end{table}

 \begin{figure}[t]
\centering
\centerline{\includegraphics[width=6.8cm]{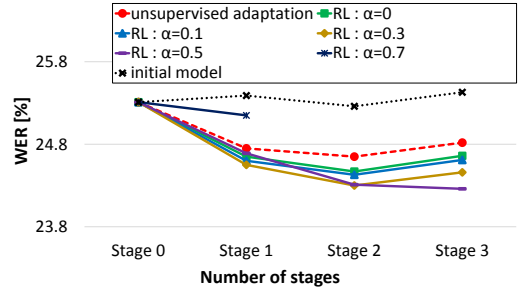}}
\vspace{-4mm}
%%\caption{Number of epochs and WERs of the unlabeled training data for the CSJ task.
%%``RL'' indicates reinforcement learning.} 
\caption{Number of stages and WERs of the large batch data.
At stage k, the RLk model is used to decode large batch \#k+1.}
%``RL'' indicates reinforcement learning.}
\vspace{-2mm}
\label{fig:csj-train}
\end{figure}

\begin{figure}[t]
\centering
\centerline{\includegraphics[width=6.8cm]{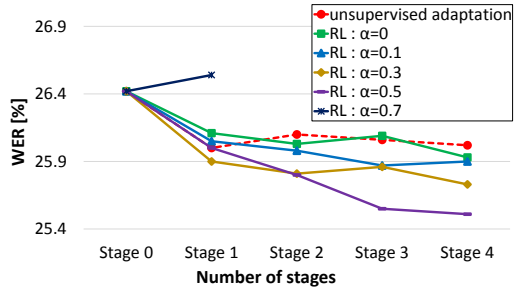}}
\vspace{-4mm}
\caption{Number of stages and WERs of the evaluation set.}
%%\caption{Number of stages and WERs of the evaluation set for the CSJ task.}  
\label{fig:csj-eval}
\vspace{-4mm}
\end{figure}

We performed the experiments using data from the Corpus of Spontaneous Japanese (CSJ)~\cite{CSJ},
and based on the CSJ recipe\footnote{\scriptsize\url{https://github.com/kaldi-asr/kaldi/tree/master/egs/csj}}
in the Kaldi speech recognition toolkit~\cite{povey2011kaldi}.
%We performed the experiments using the Corpus of Spontaneous Japanese (CSJ)~\cite{CSJ}.
%The recognition system and each parameter were based on 
%CSJ recipe\footnote{https://github.com/kaldi-asr/kaldi/tree/master/egs/csj}
%in Kaldi toolkit.
%%We performed two sets of experiments using the Corpus of Spontaneous Japanese (CSJ)~\cite{CSJ} 
%%and the Switchboard corpus (SWBD)~\cite{swbd}.
In our experiments, we made two subsets from the original CSJ training data.
The first subset contained 10 hours of data, and it was used as a labeled training set to train an initial baseline system.
The other subset had 200 hours in total, and it was further divided into
four subsets, each of which contained 50 hours of data.
These four subsets were used as the unlabeled large batches for the reinforcement learning assuming
the corresponding transcripts were not given to the system.
Additionally, the standard evaluation set of CSJ, including two hours of speech data was used to evaluate
the updated models using the same data set.
Table~\ref{dataset} summarizes these data sets.
The feedback from the users was simulated by evaluating the word error rates (WER)
of the hypotheses from the system using the reference labels, and then performing the
hypothesis selection based on the true WER.
To simulate selection errors caused by users, experiments that introduced
random swapping of the selected and unselected hypotheses were performed.

The input acoustic features for the DNN were 40 dimensional fMLLR features.
%%~\cite{fMLLR, Povey_fmllr_2006}
%%obtained from mean- and variance-normalized 13-dimensional MFCC features.
%%Before they were input to the DNN, they were
%%spliced by +/- 4 frames and 
%%transformed into a 40-dimensional feature
%%using linear discriminant analysis (LDA) and maximum likelihood linear transform (MLLT).
%%Furthermore, speaker adaptive training (SAT) was applied
%%using feature-space maximum likelihood linear regression (fMLLR)~\cite{fMLLR, Povey_fmllr_2006}.
They were computed using lattices, 
where the lattices were made by forced aligning the true labels for the training set,
and by decoding the speech data for the large batches and for the evaluation set. 
%The fMLLR was estimated from decoded lattices for the unlabeled training data.
The size of the input layer of the DNN was 1400 (spliced by +/- 17 frames).
%%The size of the input layer of the DNN was 1400 (spliced by +/- 17 frames) for CSJ
%%and 440 (spliced by +/- 5 frames) for SWBD.
%
%Different from CSJ recipe, the DNN had 812 units for the output layer.
The DNNs had 6 hidden layers with a sigmoid activation function.
They had 1905 units per hidden layer and 812 units for the output softmax layer.
%%The DNN for CSJ had 1905 units per hidden layer and 812 units for the output layer,
%%and that for SWBD had 2048 units per hidden layer and 823 units for the output layer.
%The numbers of units in the output layers correspond to the number of HMM states.

The DNN-HMM of the baseline system 
was trained by pre-training and fine-tuning using the 10-hour labeled training data.
%Mini-batch stochastic gradient descent (SGD) was used with the cross-entropy error measure.
%The mini-batch size was 256.
For the large batch based reinforcement learning,
the initial learning rates for the batches were set to 0.004, 0.002, 0.001 and 0.0005
for stages 1, 2, 3, and 4, respectively.
%%The initial learning rate was respectively set to 0.004, 0.002, 0.001 and 0.0005 in stage 1, 2, 3 and 4 for CSJ,
%%and that was set to 0.004, 0.002, 0.0005 in stage 1, 2 and 3 for SWBD.
The 10-hour labeled training data was always used by mixing it with the unlabeled large batches.
The learning rate controls for the training data set and for the large batches
were based on cross-validation using 10\% of the labeled trained data as the held out set.
The learning rate was halved when the improvement in cross-entropy on a cross-validation set
fell below 1\% in an epoch.
The upper limit of the number of iterations in each epoch was set to 7.
%For the cross-validation, 10\% of the labeled trained data were used.
%The language model was a word tri-gram.

Figure~\ref{fig:process} shows the outline of the reinforcement learning process.
The unlabeled large batch \#1 was decoded using the initial baseline DNN-HMM model.
%We first decoded the unlabeled training data using the baseline DNN-HMM as an initial model.
The Candidate 1 hypotheses were the best results in the N-best list,
and the Candidate 2 hypotheses were either 10th or fifth results in the list.
%As the rank 1 and rank 2 hypotheses, the 1st and 10th hypotheses in an N-best list
The N-best list was created from a decoded lattice.
% After that, the best text is regarded as rank 1, and the worst text as rank 2.
%We set the reward $r=1$ when the rank 1 hypothesis actually had higher 
%word accuracy than the rank 2 hypothesis, and $r=0$ otherwise.
%To simulate imperfect selection, we investigated the relation between WER of selected hypothesis 
%ad the rate of selection error by swapping the rewards value randomly.
%\ts{describe the results in the RESULTS section}
%Figure \ref{fig:error_rate} shows the results.
%When the rate of selection error was 25\%, 
%WER of selected hypothesis was higher than the rank 1 hypothesis.
%Therefore, it is considered that the hypothesis selection is effective 
%when the rate of selection error is lower than 25\%.
%We additionally tested a condition
%where 15\% of the reward values were randomly swapped for reinforcement learning. 
%To simulate imperfect selection, we additionally tested a condition
%where 15\% of the reward values were randomly swapped.
After the first updated model (RL1) was made using large batch \#1,
it was used to recognize large batch \#2.
Based on the recognition results, the model was updated, making the next model (RL2).
This process was repeated for all the large batches. 
%After an epoch of the reinforcement learning, the DNN-HMM was updated
%and used as an initial model for the next epoch.
%When updating the DNN-HMM, the labeled data set with the manual transcript
%and the unlabeled data set with the decoded label were jointly used.
For comparison purposes, unsupervised adaptation was performed,
where the model was updated using the Candidate 1 hypothesis without the hypothesis selection.

\section{Results}
\label{sec:results}
\vspace{-1mm}
%
% -----SHINOT-------THIS-IS-SWBD-----
\begin{figure}[t]
\centering
\centerline{\includegraphics[width=6.8cm]{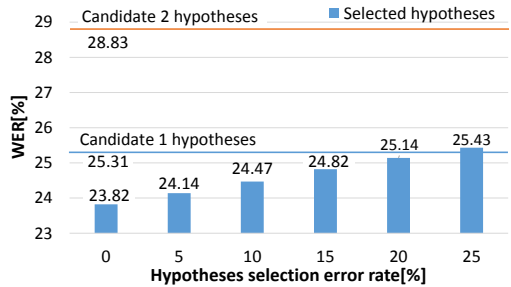}}
\vspace{-4mm}
\caption{Relationship between hypotheses selection error rate
  and WER of the selected hypotheses.}
  \vspace{-4mm}
\label{fig:error_rate}
\end{figure}

\vspace{-2mm}
% -----SHINOT-------THIS-IS-SWBD-----

\begin{figure}[t]
\centering
\centerline{\includegraphics[width=6.8cm]{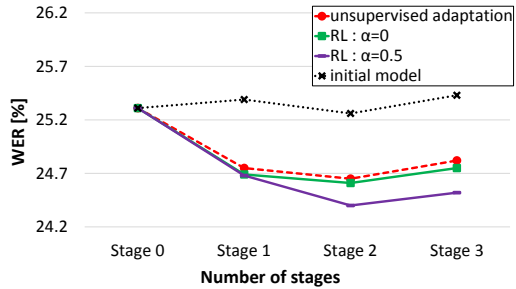}}
\vspace{-3mm}
%%\caption{Number of epochs and WERs of the unlabeled training data for the CSJ task,
%%where 15\% of the rewards were swapped to simulate selection errors.} 
\caption{Number of stages and WERs of the large batches
when there is 15\% hypotheses selection error.}
\label{fig:csj-train-15}
\vspace{-3mm}
\end{figure}
%
%%Figure \ref{fig:csj-train} shows the number of epochs and WERs of the unlabeled training data for the CSJ task 
Figure \ref{fig:csj-train} shows the WER of the successively updated models based on 
unsupervised adaptation and reinforcement learning using the large batches sequentially.
%The WER by the initial baseline model was 25.31\%.
At stage 0, the initial baseline model is used to decode large batch \#1.
The hypothesis selection is only for model update, and the WERs in the figure are all based on the 1-best result.
Therefore, differences of the WERs arise from Stage 1.
In the figure, "initial model" indicates the WERs of the large batches using the baseline initial model.
The unsupervised adaptation gave better results than the non-updated initial model.
%For the unsupervised adaptation using the rank 1 hypothesis, it is seen that lower WER was obtained than the initial model.
For the reinforcement learning, 10th-best results were used as the Candidate 2 hypotheses.
By using reinforcement learning, a larger improvement than the
unsupervised adaptation was obtained when the coefficient $\alpha$ was chosen from 0.0 to 0.5.
Choosing $\alpha$ greater than $0$ means both of the hypotheses were used.
The lowest WER was obtained when $\alpha$ was 0.5.
%% =======
%% For the unsupervised adaptation using the rank 1 hypothesis, it is seen that lower WER was obtained than the initial model.
%% As for the reinforcement learning, larger improvement than the
%% unsupervised adaptation was obtained when the coefficient $\alpha$ was chosen to less than or equal to 0.5.
%% Choosing $\alpha=0$ means using only the selected hypothesis, and choosing $\alpha$ greater than $0$ means
%% using both of the two hypotheses.
%% The lowest WER was obtained in the latter case where $\alpha$ was 0.5.
%% >>>>>>> 8cac57cc24ac2af3691de2769ada06a2409f194f
%The difference of WER between unsupervised adaptation and reinforcement learning was increased for the number of epochs.
When $\alpha$ was larger than 0.5, the second hypothesis affected the gradient too much and
WER greatly increased.
%it resulted in a large increase in the WER.
At stage 3, WER slightly increased except when $\alpha=0.5$, including the unsupervised adaptation.
%The WER gradually increased in epoch 3 in both unsupervised adaptation 
%and reinforcement learning when $\alpha$ was chosen to less than 0.5.
%The same tendencies were observed in the initial model
This was partly because our learning rate reducing strategy was not optimal, and partly because
the fourth batch simply contained relatively difficult utterances to recognize, as it is seen that
the WER using the initial model was also higher compared to the other large batches.
%We conjecture this was because of overtraining to the labeled training set.

To evaluate the updated models using the same data set, 
Figure \ref{fig:csj-eval} shows WERs of the common evaluation set.
The WER by the initial baseline model was 26.42\%, and 
the unsupervised adaptation gave 0.4\% absolute improvement at the 4th stage.
%The reinforcement learning gave improvements when $\alpha$ was less than or equal to 0.5. 
Consistent improvement was observed by the reinforcement
learning with $\alpha=0.5$, and it gave the lowest WER of 25.51\% at the 4th stage.

Figure \ref{fig:error_rate} shows the simulated results of the relation between the selection error rate
by the users and the WERs of the selected hypotheses.
When the selection error rate is equal to or lower than 20\%, we can expect lower WER in the selected hypotheses
than the Candidate 1 hypotheses.
Based on this analysis, we next investigated the performance of the reinforcement learning
% -----SHINOT-------THIS-IS-SWBD-----
%We next investigated the performance of the reinforcement learning
% -----SHINOT-------THIS-IS-SWBD-----
when there were 15\% errors in the hypotheses selection.
Figure \ref{fig:csj-train-15} shows the WERs. %  of the unlabeled training data
The WER of stage 0 is the same as that of the Figure \ref{fig:csj-train}.
It is confirmed that the reinforcement learning still outperformed the unsupervised adaptation.
At the 3rd stage, a slight increase in WER was observed both for the unsupervised adaptation
and the reinforcement learning due to the same reason as before.

%Figure~\ref{fig:csj2} shows the number of epochs and WERs of the unlabeled training data of the CSJ task
%when 15\% of the rewards were randomly swapped to simulate selection errors by the users. % for reinforcement learning.
%Similar to the results in Figure \ref{fig:csj1}, the reinforcement learning showed lower WER than the unsupervised adaptation.
%and it decreases in the order of unsupervised adaptation, $\alpha=0$, $\alpha=0.3$.
%Compared Figure \ref{fig:swbd1} with Figure \ref{fig:swbd2} at corresponding $\alpha$, 
%we can see WER of hypothesis for unlabeled data becomes worse when 15\% of the rewards were swapped.
%Table \ref{CSJ-result-15} shows the WERs on the evaluation set. 
%In this experiment,
%The baseline is the same as in the Table \ref{CSJ-result}.
%It can be seen the reinforcement learning still outperformed unsupervised adaptation
%when $\alpha$ is 0.5. 
%Therefore, it is assumed that our proposed reinforcement learning is effective in an actual environment.

\begin{figure}[t]
\centering
\centerline{\includegraphics[width=6.8cm]{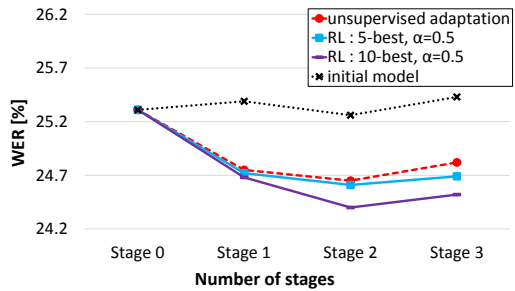}}
\vspace{-3mm}
\caption{Number of stages and WERs of the large batches when
  the 5th and 10th best hypotheses were used as the Candidate 2 results.
  15\% selection error rate is simulated.}
%when there is 15\% of the rewards were swapped to simulate selection errors.
%$\alpha$ is set to 0.5 in reinforcement learning using 5th or 10th hypothesis as rank 2 hypothesis.}
%%\caption{Number of epochs and WERs of the unlabeled training data for the CSJ task, 
%%where 15\% of the rewards were swapped to simulate selection errors.
%%$\alpha$ is set to 0.5 in reinforcement learning using 5th or 10th hypothesis as rank 2 hypothesis.}
\vspace{-4mm}
\label{fig:csj-train-nbest}
\end{figure}

Finally, we have evaluated the performance of the reinforcement learning when
the 5th-best results were used as the Candidate 2 hypotheses instead of the 10th-best results.
%In our experiments, 10th hypothesis in the N-best list was used as rank 2 hypothesis. 
%To investigate an effect of quality of rank 2 hypothesis, 
%we performed reinforcement learning using 5th hypothesis in the N-best list as rank 2 hypothesis. 
Figure \ref{fig:csj-train-nbest} shows the WERs with 15\% selection errors. % of the unlabeled training set, 
%when 15\% of the rewards were randomly swapped.
For reinforcement learning, $\alpha$ was set to 0.5.
While the improvement became small, reinforcement learning still gave better results than the
unsupervised adaptation.

\section{Conclusion}
\label{sec:conclusion}
\vspace{-1mm}
In this paper, we have proposed a policy gradient-based reinforcement learning framework for speech recognition systems,
and also have proposed a hypothesis selecting-based reinforcement learning method as
a particular instance of the framework.
%the best recognition hypothesis among two hypotheses.
%We have presented considerations about the connection of our proposed method to
%the general MDP based formulation of reinforcement learning,
%and have shown that out method fits in the framework.
In the experiments, we have shown that the proposed method reduces WER compared to the unsupervised adaptation.
%by using both of the two hypotheses.
The tendencies were the same when 15\% of simulated noise in the hypothesis selection was introduced,
while the improvement became slightly smaller.
%%The same tendencies were observed in the two experiments using the Japanese CSJ and English SWBD datasets,
%%and in the experiment where 15\% of simulated noise in the hypothesis selection was introduced.
When the number of stages was increased, there was a tendency for the WER to increase 
in both the unsupervised adaptation and the reinforcement learning in several cases.  
Future work includes addressing the problem of overtraining by adjusting the strategy for the learning rate 
and the number of iterations in each stage, and improving the performance by investigating more effective ways to update the model.

%%When the number of epochs was increased, there was a tendency that the evaluation set WER increased while
%%the training set WER decreased both in the unsupervised adaptation and the reinforcement learning.
%%Future work includes addressing the problem of the overtraining.
%%For this, the DNN re-estimation process should be improved.
%%Another point is that, in the experiments, we repeatedly used the same data set in the iteration of the epochs.
%%However, this was not realistic in the assumed story to service a vast number of users on the internet.
%%At the same time, this might have been the reason for the overtraining.
%%For the assumed scenario, we can use and discard plentiful input utterances without keeping it,
%%which may alleviate the problem of the overtraining.

% References should be produced using the bibtex program from suitable
% BiBTeX files (here: strings, refs, manuals). The IEEEbib.bst bibliography
% style file from IEEE produces unsorted bibliography list.
% -------------------------------------------------------------------------

\bibliographystyle{IEEEbib}
%\bibliography{strings,refs}

\end{document}